\documentclass[conference]{IEEEtran}
\IEEEoverridecommandlockouts
\usepackage{cite}
\usepackage{amsmath,amssymb,amsfonts}
\usepackage{algorithmic}
\usepackage{graphicx}
\usepackage{textcomp}
\usepackage{xcolor}
\usepackage{subfig}

\usepackage{hyperref}
\hypersetup{colorlinks,linkcolor=blue,citecolor=blue,urlcolor=magenta, linktocpage,plainpages=false, bookmarksopen = true}

\def\BibTeX{{\rm B\kern-.05em{\sc i\kern-.025em b}\kern-.08em
    T\kern-.1667em\lower.7ex\hbox{E}\kern-.125emX}}
\usepackage{bbold}
\begin{document}

\title{Understanding the Ability of Deep Neural Networks to Count Connected Components in Images\\
}

\author{\IEEEauthorblockN{Shuyue Guan}
\IEEEauthorblockA{\textit{Department of Biomedical Engineering} \\
\textit{The George Washington University Medical Center}\\
Washington DC, USA \\
frankshuyueguan@gwu.edu \\ https://orcid.org/0000-0002-3779-9368}
\and
\IEEEauthorblockN{Murray Loew}
\IEEEauthorblockA{\textit{Department of Biomedical Engineering} \\
\textit{The George Washington University Medical Center}\\
Washington DC, USA \\
loew@gwu.edu}
}
\IEEEoverridecommandlockouts
\maketitle

\IEEEpubidadjcol

\begin{abstract}
Humans can count very fast by subitizing, but slow substantially as the number of objects increases. Previous studies have shown a trained deep neural network (DNN) detector can count the number of objects in an amount of time that increases slowly with the number of objects. Such a phenomenon suggests the subitizing ability of DNNs, and unlike humans, it works equally well for large numbers. Many existing studies have successfully applied DNNs to object counting, but few studies have studied the subitizing ability of DNNs and its interpretation.
In this paper, we found DNNs do not have the ability to generally count connected components. We provided experiments to support our conclusions and explanations to understand the results and phenomena of these experiments. We proposed three ML-learnable characteristics to verify learnable problems for ML models, such as DNNs, and explain why DNNs work for specific counting problems but cannot generally count connected components.
\end{abstract}


\section{Introduction}
Humans detect the number of objects by a spontaneous “number sense” (called “subitizing”) if the number is small, and by deliberately counting and memorizing for a large number. Humans can count very fast by subitizing, but slow substantially as the number of objects increases. In deep learning, previous studies have successfully applied deep neural networks (DNN) to object counting. A trained DNN detector can count the number of objects in an amount of time that increases slowly with the number of objects. Such a phenomenon suggests the subitizing ability of DNNs, and unlike humans, it works equally well for large numbers. Many existing studies have examined the applications of DNNs to object counting, but few studies have studied the subitizing ability of a DNN and its interpretation. In a DNN, subitizing does not first detect and then count objects (as in traditional image processing), but it recognizes the number of objects in an image \textit{directly}. The DNN’s ability to transform images to binary images through segmentation has been verified by numerous studies. For a DNN’s subitizing, however, its essential ability is to detect connected components in binary images because object counting depends on their connectedness. In this study, we have created binary images containing various numbers of connected components and tested the ability of DNNs to count the number of pixels and connected components in the images. Each component has its size, shape, or/and location selected at random.

\subsection{Our works}
A deep-neural network pixel-wise counter was trained on binary images containing numbers of individual object pixels (\textit{i.e}., pixel with value 1 is the object and 0 is background) spanning the entire range from 1000 to 3000 and tested on images that included the range from 3 to 10000 pixels. The experiment was repeated 10 times, yielding an overall error rate of (0.094 $\pm$ 0.002)\%. In addition, we interpreted this pixel-wise counting ability by comparing different structures and analyzing weights and data flow within neural networks. Understanding this ability is an important step toward explaining and trusting object counting with deep learning.

Unfortunately, our experiments show that DNNs cannot generally count connected components in the image. Then, we tentatively provided three explanations based on \textit{learnability} to understand why DNNs cannot \textit{generally} count objects but successfully work for many \textit{specific} counting tasks. We consider that, for machine learning (ML) models (like DNNs), a learnable problem should have the three characteristics (ML-learnable characteristics):
\begin{enumerate}
    \item It has a \textbf{finite domain}.
    \item It has enough number of data to show its pattern (\textbf{enough data}).
    \item Its subsets have the similar pattern as the whole set (\textbf{pattern consistency}).
\end{enumerate}
The first two characteristics are simple and well-known but the third one is profound and needs further studies and discussions.

In summary, around using DNNs to count connected components in image, we have done several experiments and tried to provide some insights into the DNN and data for machine learning. This study may raise other questions and discover the starting points of ways for future researchers to make progress in the understanding of deep learning.

\subsection{Related works}
In deep learning, previous studies have successfully applied DNNs to object counting \cite{ilyas_convolutional-neural_2019,aich_object_2019,liu_crowd_2018,yao_deep_2017,he_delving_2017}. For the leaf counting challenge \cite{noauthor_leaf_nodate}, leaves are firstly segmented by the SegNet \cite{badrinarayanan_segnet:_2017} to obtain a binary leaf image and then the number of leaves is counted by using the breadth first search (BFS) or regression. It is counting by \textit{segmentation}. Object counting is also done by \textit{recognition} and \textit{classification}. Object counting tasks can be considered as recognition problems. Oñoro-Rubio \& López-Sastre (2016) \cite{onoro-rubio_towards_2016} applied the CNN model to transform an image with vehicles to a density map. Then, the number of vehicles is predicted by a regression model from the density (heat) map. The work of Zhang et al. (2015) \cite{zhang_salient_2015} seems similar to ours but this work only classifies images with objects into 5 categories: 0 (no clear object), 1, 2, 3, and 4+ (four and more objects). The main drawback of counting through classification is that the countable number of objects is known and limited. 

In general, the goal of automated object counting is to count any number of objects (more than the numbers for training) \textit{directly} from images. The direction means the input of model is an image and its output is the object number. In this study, we examine a simplified situation; the questions are: can DNNs count the \textit{general} connected components in images? Why or why not? The generalization means the connected components could be in any shapes, sizes, and positions. Nasr et al. (2019) \cite{nasr_number_2019} shows that in the final layer of a trained (on ImageNet) CNN model, 9.6\% (3,601/37,632) units can react to the number of objects (called numerosity-selective units). By using the responses of these units can build tuning curves to estimate the number of connected components (range from 1 to 30) in images. The problem of this paper is that the numerosity-selective units are selected and tuned by other methods and experiments; the processes are manually interfered and thus the counting is neither direct nor automatic.
\section{Count pixels}
The first part of experiments is to examine whether DNNs can count the number of pixels:
\begin{itemize}
    \item \textbf{Dataset}: 256x256 binary images that the label of background is “0” and objects is “1” (see Fig. \ref{fig:3}, Random pixels). The training set contains 10,000 images with 1000-3000 “1” pixels (20\% for Validation) and test set contains 1,000 images with 1-10000 “1” pixels. All “1” pixels are randomly located.
    \item \textbf{Models}: a fully-connected neural network (FCNN) without hidden layers (Perceptron, $M_0$) and a FCNN with only one (128 neurons) hidden layer ($M_1$).
\end{itemize}

\begin{figure}[htbp]
\centerline{\includegraphics[width=0.48\textwidth]{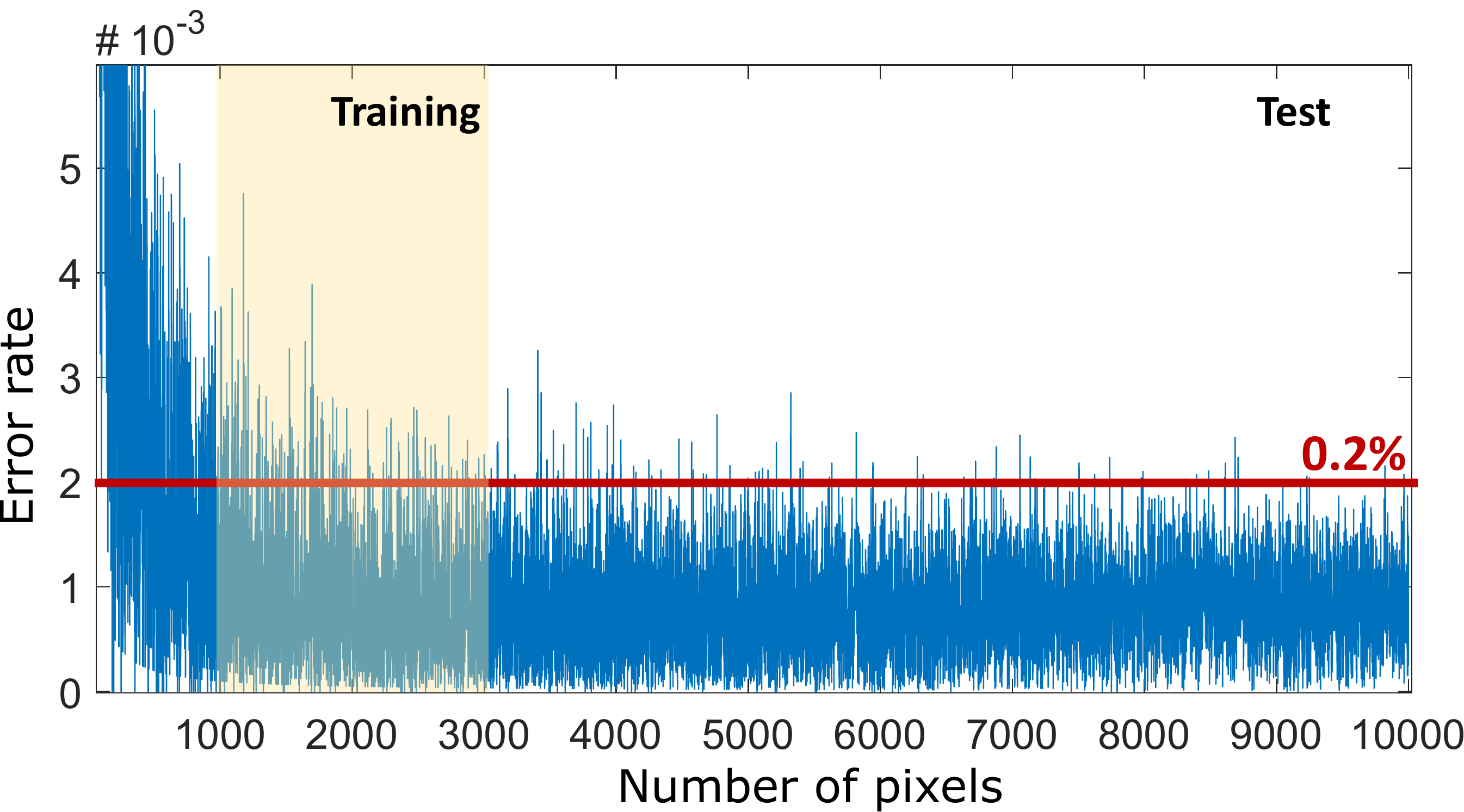}}
\caption{Test results of $M_0$. The error rate is $|predicted\ number-real\ number|\div (real\ number)$ on test set. Yellow box shows the range of training set. Red line is at 0.2\% error rate.}
\label{fig:1}
\end{figure}

\begin{figure}[htbp]
\centerline{\includegraphics[width=0.48\textwidth]{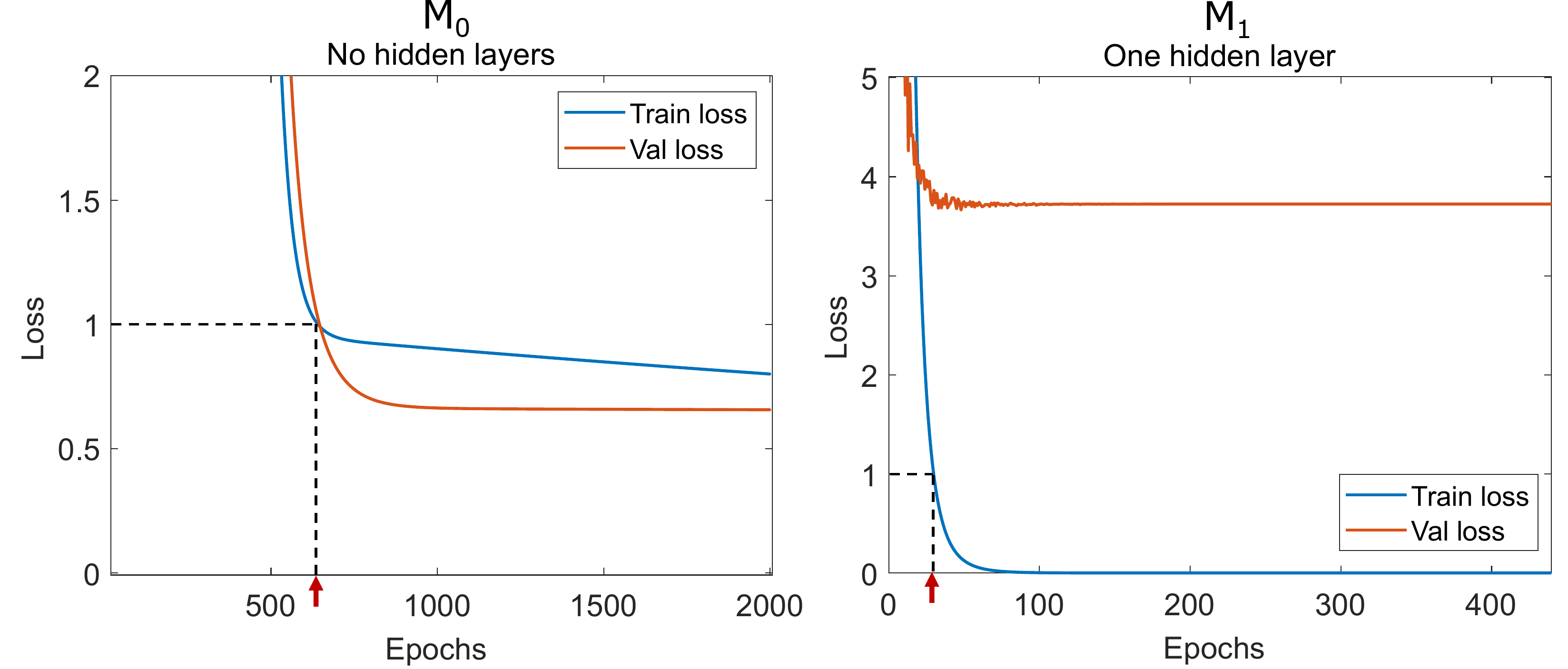}}
\caption{Training processes of $M_0$ and $M_1$. Note: the two figures have different scales. Red arrows show epochs where the train loss is at 1.0. }
\label{fig:2}
\end{figure}

Naturally, the results agree with our expectations. The experiment was repeated 10 times, yielding an overall error rate of (0.094 $\pm$ 0.002)\%. As Fig. \ref{fig:1} shown, a simple deep learning model can learn the counting of pixels. And we found that the training process of $M_1$ was much faster than that of $M_0$. As shown in Fig. \ref{fig:2}, for $M_0$, to reach 1.0 train loss needs more than 500 epochs but $M_1$ needs less than 50 epochs. This phenomenon is anti-intuitive and interesting because $M_1$ is a more complex model than $M_0$ and $M_1$ is expected to spend more times to train. In summary, \hypertarget{ques}{two questions} derive from this experiment: 1) how does the DNN count the number of pixels? 2) why is the model with more hidden layers trained faster?

\subsection{Explanation}
We suppose that $X$ is the vector of a flattened binary image whose labels for background pixels are “0” and objects are “1”:
\[
X=\left( \begin{array}{c} x_1 \\ x_2 \\ \vdots \\ x_N \end{array} \right) \quad x_i \in \{0,1\}
\]
The simplest way to obtain the number of objects is to multiply a one vector $\mathbb{1}$ by the $X$:
\[
\mathbb{1}X=\left(1\ 1\ \ldots \ 1\right)\left( \begin{array}{c} x_1 \\ x_2 \\ \vdots \\ x_N \end{array} \right)
\]
Hence, if the $M_0$ model has learned to count, we have:
\[
AX+B=\mathbb{1}X
\]
Where A is the weights matrix and B is the bias of $M_0$ model. Obviously, one correct solution is $A=\mathbb{1}$ and $B=0$. If all weights in $A$ are equal to $z_0$, the loss of training $M_0$ is $N(1-z_0)$. Therefore, the destination of training $M_0$ is:
\begin{equation}
    \label{eq:1}
    z_0=1
\end{equation}

For $M_1$ learning to count:
\[
A_2 r(A_1 X+B_1)=\mathbb{1}X
\]
Where $r(\cdot)$ is ReLU activation function: $r(x)=\max \{0,x\}$. We suppose that $B_1=0$ and all weights in $A_1$ and $A_2$ are equal to $z_1 \ (z_1>0)$, then:
\[
\sum \left( z_1 \cdot r \left(\sum \left( z_1 \cdot x_i \right) \right)\right)=\sum x_i
\]
\[
\sum \left( z_1 \cdot r \left(z_1 \sum x_i \right)\right)=\sum x_i
\]
If $X$ contains $k$ objects, then $\sum x_i=k$:
\[
\sum \left( z_1 \cdot r \left(k \cdot z_1\right)\right)=k
\]
\[
Nz_1^2=1
\]
The loss of training $M_1$ is $(1-Nz_1^2)$. Therefore, the destination of training $M_1$ is $(z_1>0)$:
\begin{equation}
    \label{eq:2}
    z_1=\frac{1}{\sqrt{N}}
\end{equation}

Since the total number of pixels $N$ is a very large value, the destination of training $M_1$ is $z_1 \approx 0$. From Eq. \ref{eq:1}, the destination of training $M_0$ is $z_1 = 1$. Whereas the fact that all weights $(z_0,\ z_1)$ are initialized near 0, the destinations of $z_1$ for training $M_1$ are much closer to their initialization than that of $z_0$ for training $M_0$. Thus, the training process of $M_1$ is much faster then that of $M_0$ (see the phenomenon in Fig. \ref{fig:2}). Eq. \ref{eq:1} and \ref{eq:2} provide answers to the \hyperlink{ques}{two questions} stated before.
\section{Count connected components}
\begin{figure}[htbp]
\centerline{\includegraphics[width=0.48\textwidth]{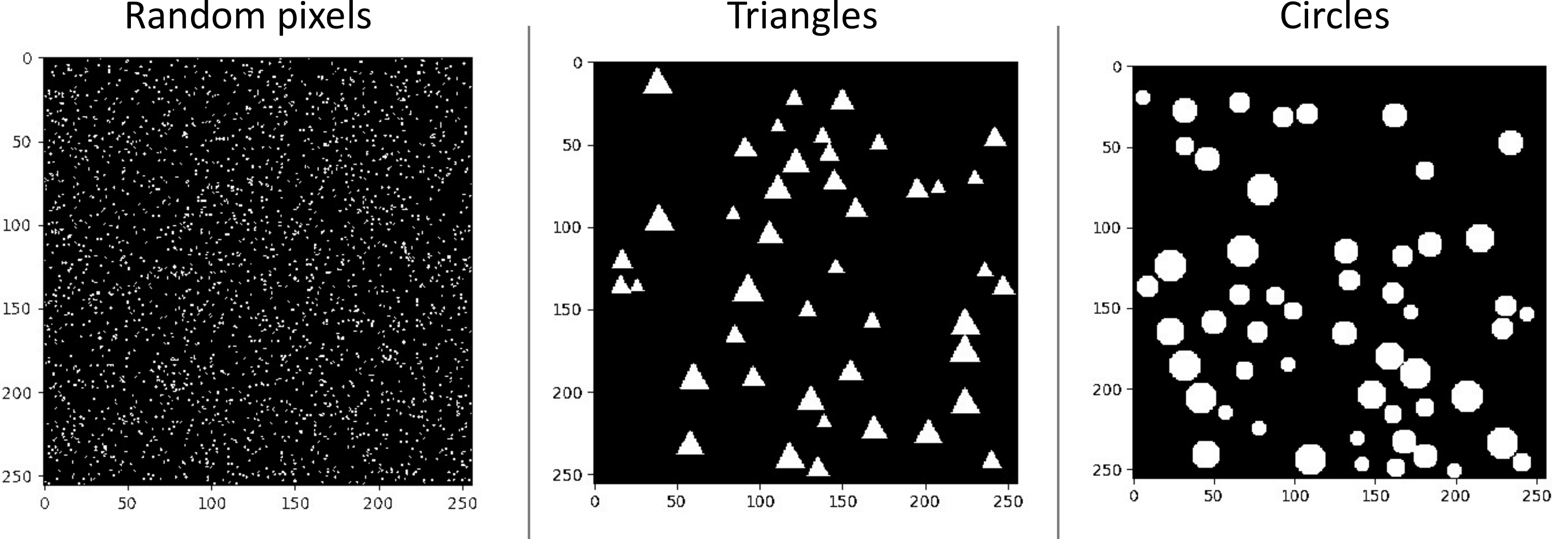}}
\caption{The three types of datasets.}
\label{fig:3}
\end{figure}

These experiments used three types of datasets and four models:
\begin{itemize}
    \item \textbf{Dataset}: 256x256 binary images with “0” and “1” pixels. These images contain various numbers of objects, which are defined by \textit{4-neighbors} connected components.
    The training set has 10,000 images (20\% for Validation) and test set has 1,000 images. Their labels are the numbers of objects (connected components). The types of these objects are (Fig. \ref{fig:3}):
        \begin{enumerate}
            \item Random pixels (same as previous experiments)
            \item Triangles (with different sizes)
            \item Circles (with different sizes)
        \end{enumerate}
        For random pixels sets, the only one parameter is the number of pixels; more pixels may form more connected components. For triangle and circle sets, there are two parameters: object size (diameter) range and object number. Objects will be generated in that size range and located randomly.
    \item \textbf{Models}:
        \begin{enumerate}
            \item $M_0$
            \item $M_1$
            \item Transfer learning; VGG16\cite{simonyan_very_2014} + $M_1$ ($M_T$)
            \item MNIST CNN \cite{noauthor_convolutional_nodate} ($M_C$)
        \end{enumerate}
\end{itemize}

\begin{figure}[htbp]
\centerline{\includegraphics[width=0.4\textwidth]{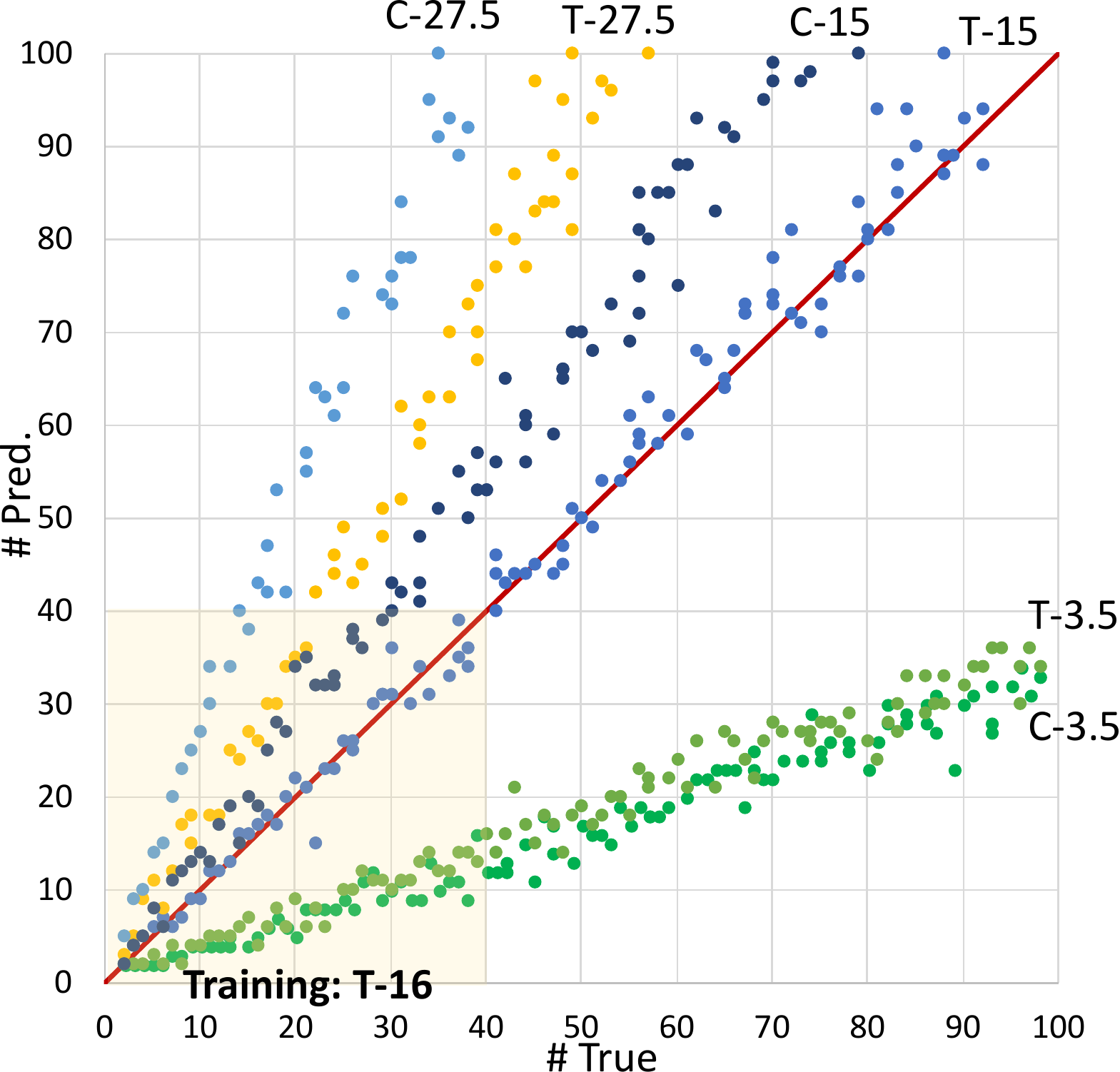}}
\caption{The predictions of $M_C$ trained by Triangle set: size $\in [2,30]$ (mean-size is 16), objects number $\in [2,40]$. T-16 means the Triangle set with 16 mean-size. Yellow box shows the range of training set. The x,y-axis are real and predicted object numbers. All test sets have the same range of objects number $\in [2,100]$ but different sizes and/or shapes. In specific, (T-15) Triangle set: size $\in [10,20]$; (T-27.5) Triangle set: size $\in [25,30]$; (T-3.5) Triangle set: size $\in [2,5]$; (C-15) Circle set: size $\in [10,20]$; (C-27.5) Circle set: size $\in [25,30]$; (C-3.5) Circle set: size $\in [2,5]$.}
\label{fig:4}
\end{figure}

By combination, there are 12 experiments. None of them, however, shows that DNNs have the ability to learn counting of connected components in the image. No matter how to design the training and test sets, once the sizes or types of objects in test sets differ from training sets, the predictions on test sets become incorrect. For example, Fig. \ref{fig:4} shows the predictions of $M_C$. Details are in the caption. Only the test set T-15 is well predicted because its average size is very close to that of the training set (T-16).

\begin{figure}[htbp]
\centerline{\includegraphics[width=0.35\textwidth]{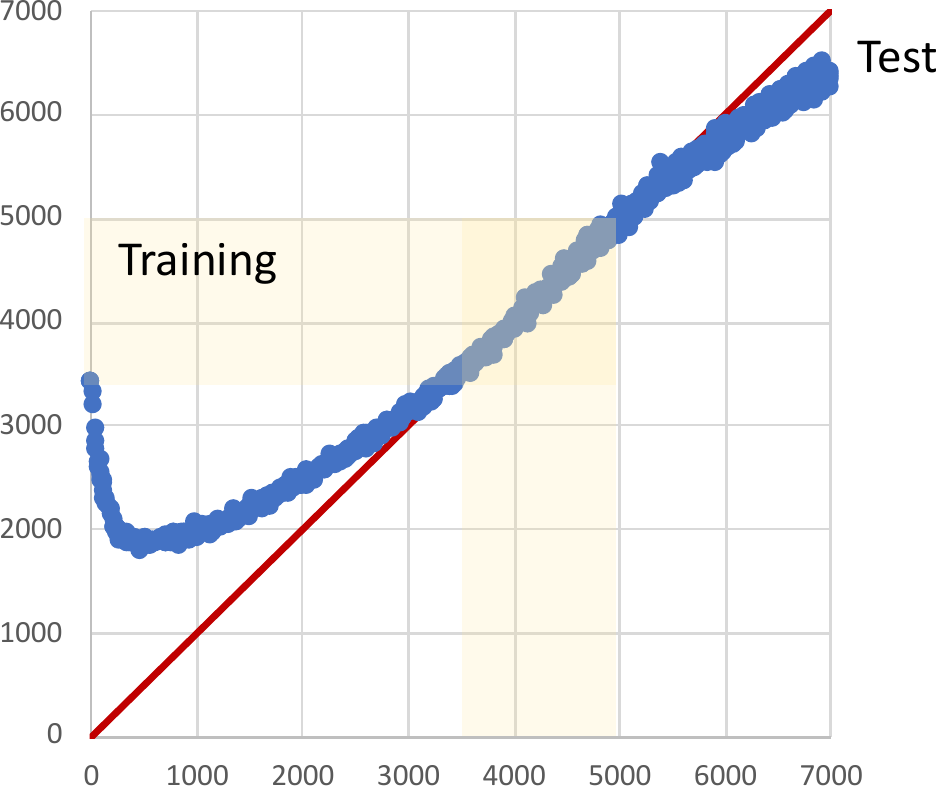}}
\caption{The predictions of $M_T$ trained by random pixels set: objects number $\in [3500,5000]$. Yellow box shows the range of training set. The x,y-axis are real and predicted object numbers. The test set has objects number $\in [10,7000]$.}
\label{fig:5}
\end{figure}

Another example in Fig. \ref{fig:5} shows the predictions of $M_T$ on random pixels sets. If and only if the object numbers in images have been included in training set, the trained $M_T$ model can perform well for such numbers in test set.

\begin{figure}[htbp]
\centerline{\includegraphics[width=0.48\textwidth]{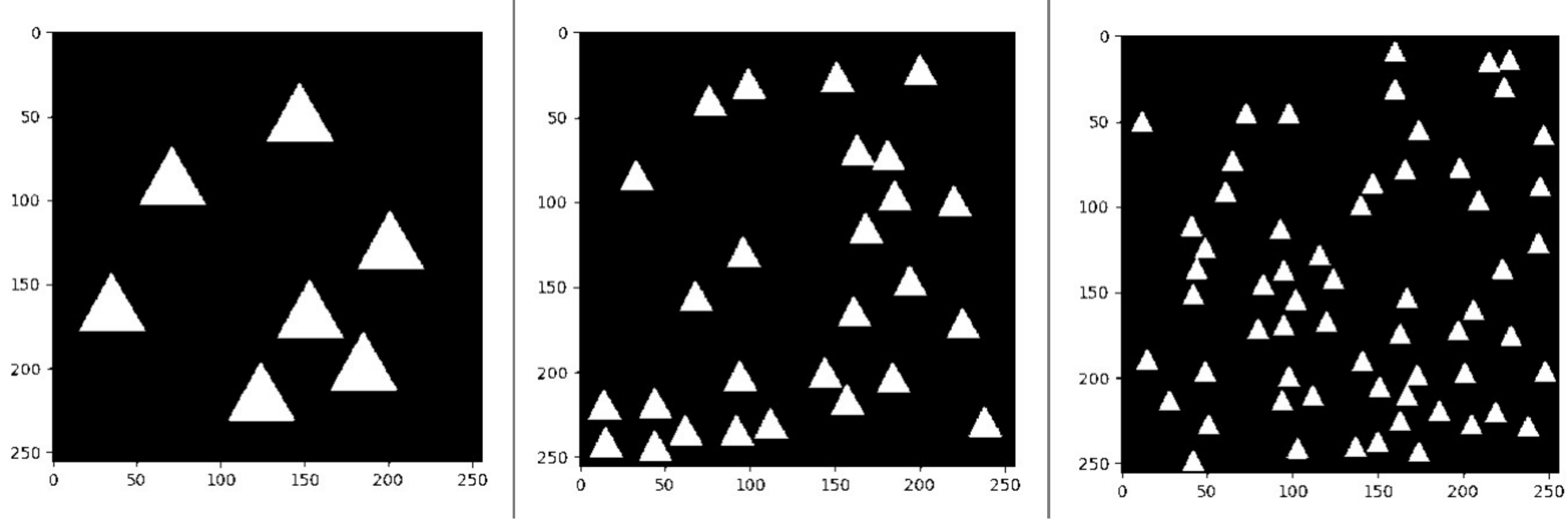}}
\caption{Examples that pixels number of all objects in an image is about 5000.}
\label{fig:6}
\end{figure}

For all datasets above, an image with more components may has more pixels because the average size of objects in each dataset is fixed. Although sizes are random in a range, the average size is stable when the number of objects increases. Thus, models could learn the  regression of the number of objects on the number of total pixels. To exclude this situation, we built additional datasets that pixels number of all objects in every image is about 5000. Fig. \ref{fig:6} shows some images of these datasets. Since pixels number of objects is fixed, the images with more objects have smaller object sizes.

\begin{figure}[htbp]
\centerline{\includegraphics[width=0.4\textwidth]{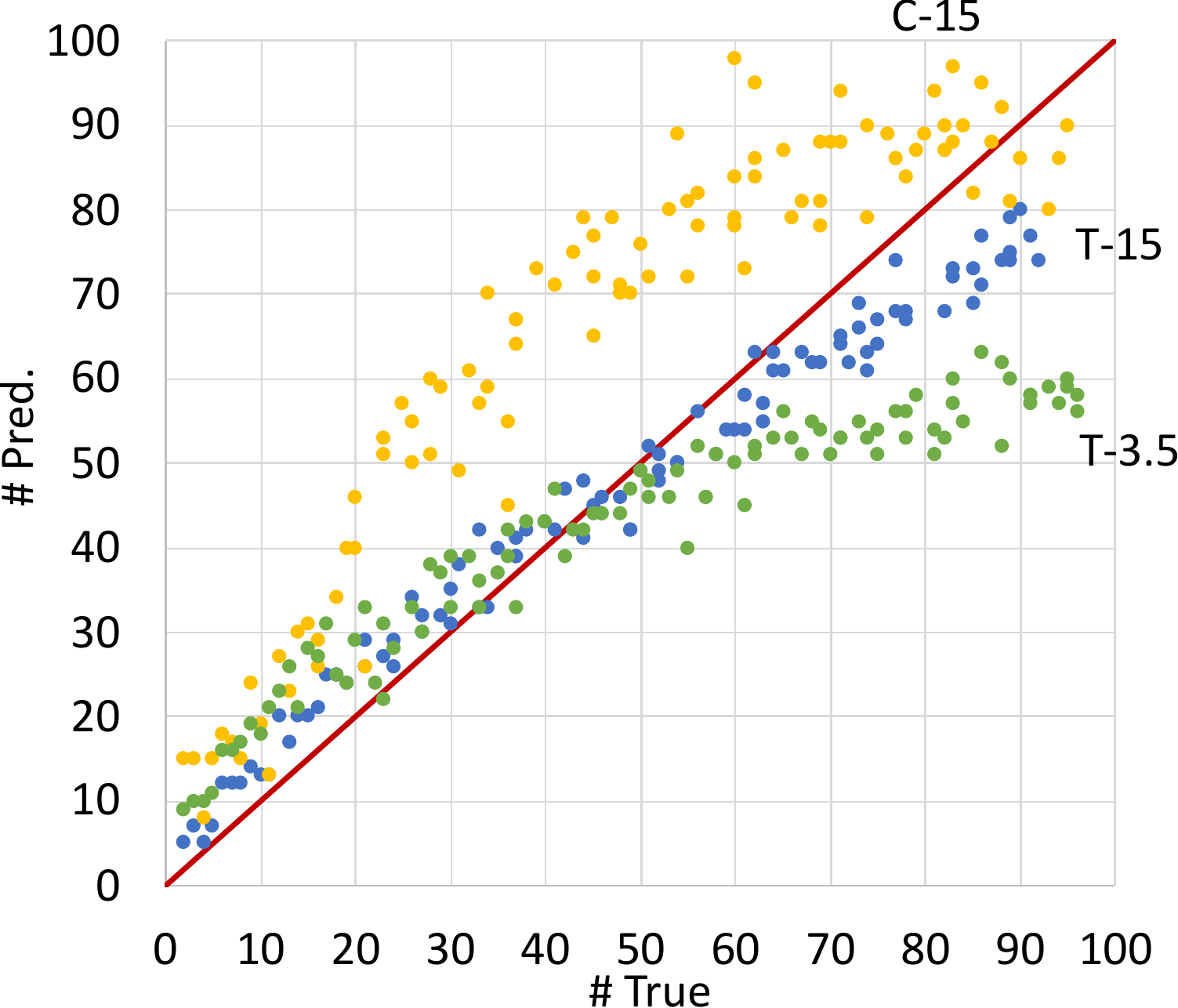}}
\caption{The predictions of $M_T$ trained by fix-object-pixel-number Triangle set: size $\in [10,50]$, objects number $\in [5,100]$. The x,y-axis are real and predicted object numbers. All test sets are fix-object-pixel-number sets and have the same range of objects number $\in [2,100]$ but different sizes and/or shapes. In specific, (T-15) Triangle set: size $\in [10,20]$; (T-3.5) Triangle set: size $\in [2,5]$; (C-15) Circle set: size $\in [10,20]$.}
\label{fig:7}
\end{figure}

Even for such fix-object-pixel-number datasets, the predictions on test sets are still incorrect. For example, Fig. \ref{fig:7} shows the predictions of $M_T$. Details are in the caption.

In summary, these results imply that DNNs cannot count the general connected components in images. The generalization is for components in any shapes, sizes and positions. And an ideal DNN counter is like a magic box, whose inputs are raw images and outputs are the numbers of objects to be counted. Hence, there are two questions need to answer: 1) why cannot DNNs achieve general counting? 2) how do DNNs successfully work for many specific counting applications in recent studies? Next, we will provide answers based on three aspects of ML models' learnability to the two questions.

\section{ML-learnable characteristics}
For ML models (such as DNNs), a learnable problem should have the three characteristics (\hypertarget{ml-learn}{ML-learnable characteristics}):
\begin{itemize}
    \item It has a finite domain.
    \item It has enough number of data to show its pattern.
    \item Its subsets have the similar pattern as the whole set (pattern consistency).
\end{itemize}

\subsection{Finite domain}
We think all patterns of problems could be categorized in two types: \textit{generatable} and \textit{checkable}. The generatable patterns mean data with that patterns could be generated by certain rules, such as even numbers are generated by multiplying 2 and integers. And the checkable patterns mean data could be checked/verified having that patterns, but we cannot find specific rules to create such data such as the prime numbers.

For data with generatable patterns, if the generative rules have been learned (like the rule-based machine learning \cite{coulibaly_rule-based_2020,weiss_rule-based_1995}), we consider that the patterns have been learned, even if the domain is infinite. Otherwise, for the problem has infinite domain, if the ML model cannot learn its generative rules or its patterns are not generatable, this problem is not ML-learnable.

Finite/infinite domain of the problem and finite/infinite amount of data in the problem are two different things. If a problem has finite number of data, its domain must be finite; thus, it is ML-learnable because a ML model can memorize all its data, like the k-nearest neighbors (KNN) or DNNs. On the other hand, if a problem has infinite number of data, its domain could be either finite or infinite. The problem having finite domain is ML-learnable because a non-parametric ML model (\textit{e.g.}, Gaussian mixtures) can learn to cover its domain. Hence, ML-learnability is related to the domain of the problem instead of the amount of data.

\begin{figure}[htbp]
    \centering
    \subfloat[\centering Train $x\in \lbrack -10,10\rbrack$ ]{{\includegraphics[width=0.22\textwidth]{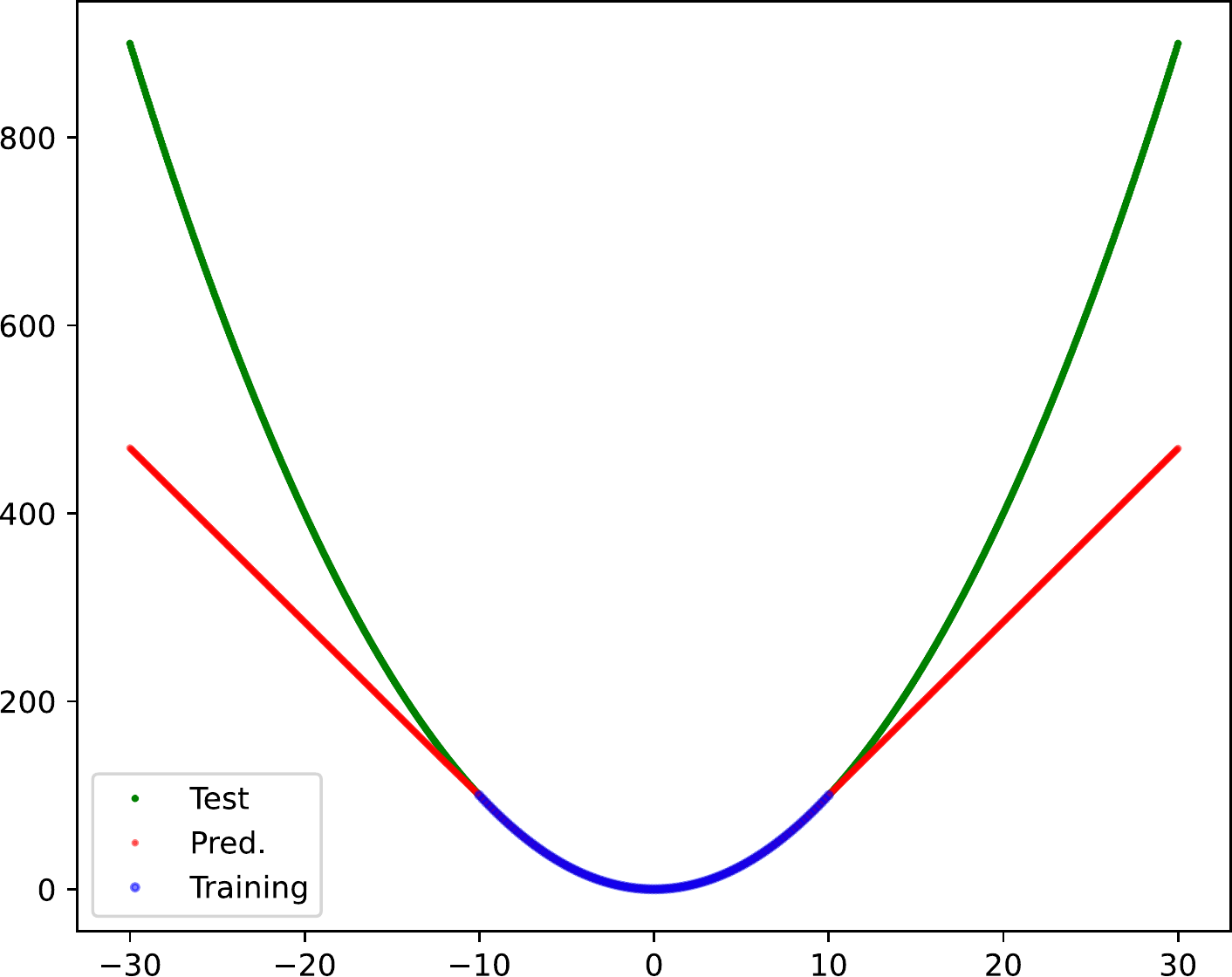} }}
    \quad
    \subfloat[\centering Train $x\in \lbrack -5,15\rbrack$ ]{{\includegraphics[width=0.2085\textwidth]{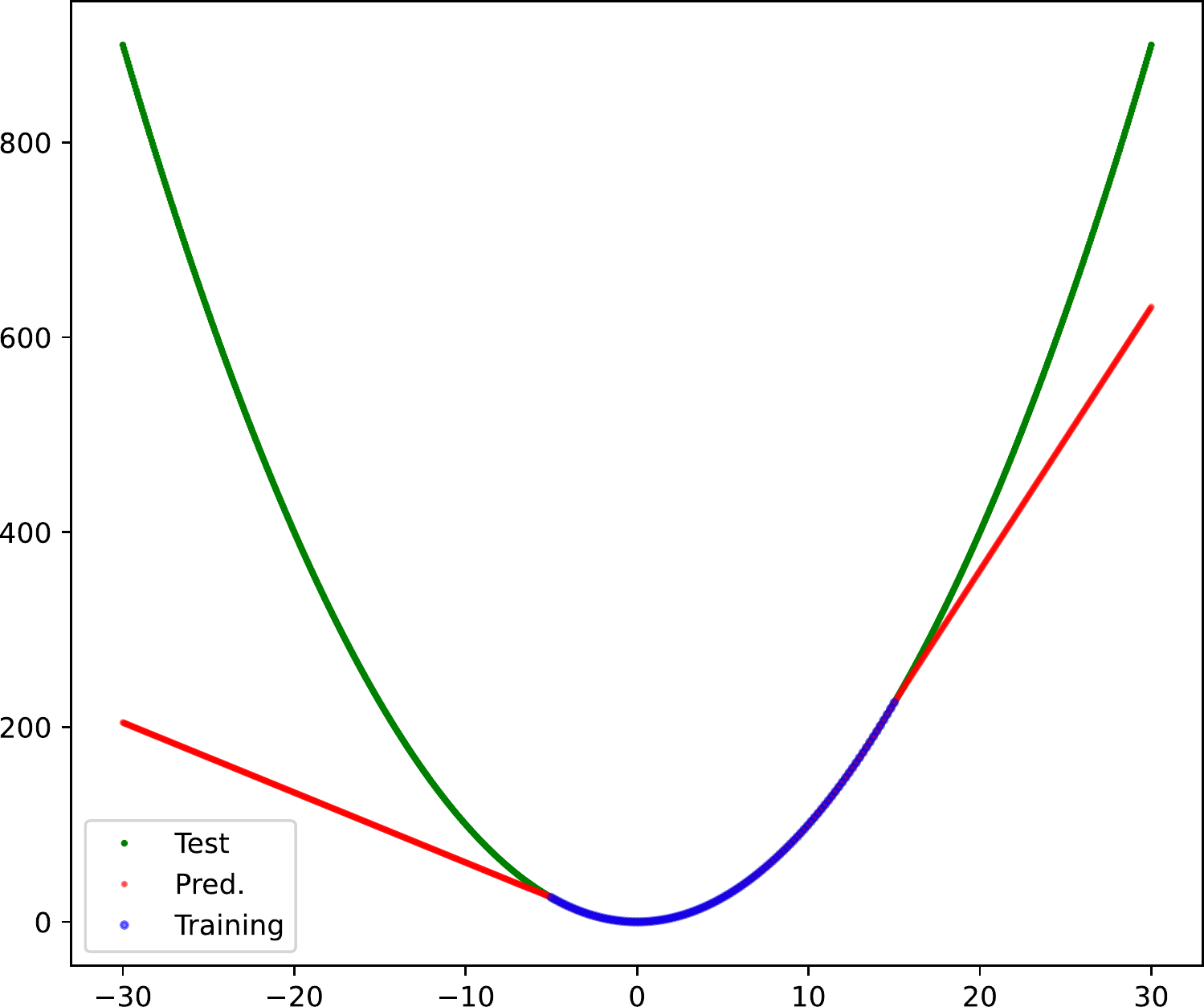} }}
    \caption{To fit $y=x^2$ by a FCNN. Blue is the training set; green is the test set $x\in \lbrack -30,30\rbrack$; red is the prediction of $y$ for $x\in \lbrack -30,30\rbrack$.}
    \label{fig:8}
\end{figure}

In summary, having finite domain is necessary to a ML-learnable problem. Although infinite domain with generatable patterns is also ML-learnable in theory, it is difficult to train a rule-based ML model to learn these patterns, and many ML models cannot learn generative rules. For example, we fitted $y=x^2$ by a FCNN. The set of $(x,y)$ has infinite domain and its pattern is generatable. As shown in Fig. \ref{fig:8}, the FCNN model does not learn the pattern because the model only predicts correctly in the same range of training.

The general tasks of counting connected components have infinite domain problems because the image sizes (inputs) and the number of components (outputs) could be arbitrary. And the pattern of such problems is not generatable; there are no specific rules to generate components because their shapes, sizes and positions are arbitrary too. Therefore, general counting connected components is not ML-learnable.

\subsection{Enough data}
\begin{figure}[htbp]
    \centering
    \subfloat[\centering 50 data points ]{{\includegraphics[width=0.3\textwidth]{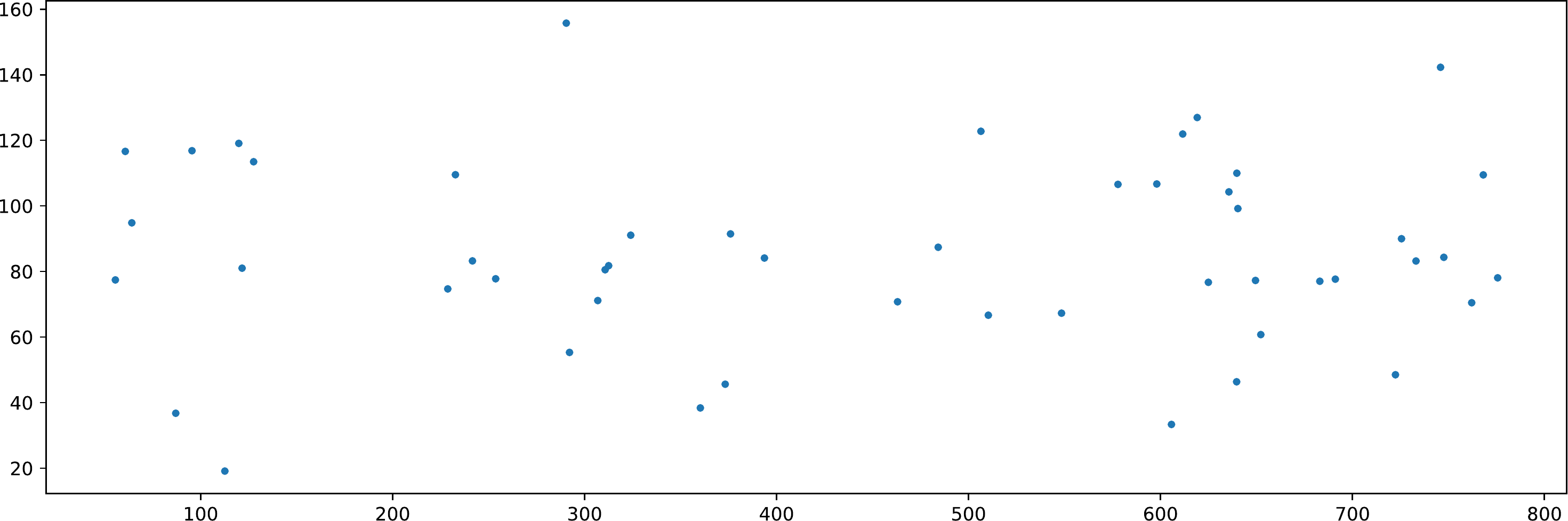} }}
    \\
    \subfloat[\centering 200 data points ]{{\includegraphics[width=0.3\textwidth]{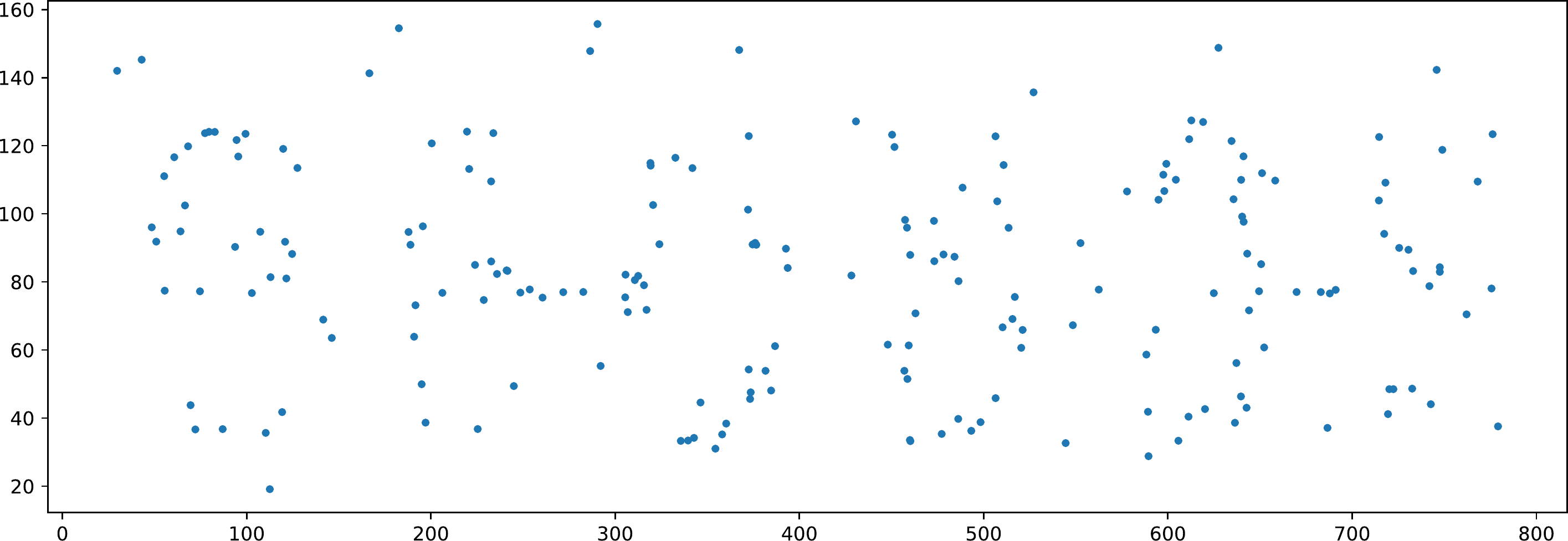} }}
    \\
    \subfloat[\centering 2000 data points ]{{\includegraphics[width=0.3\textwidth]{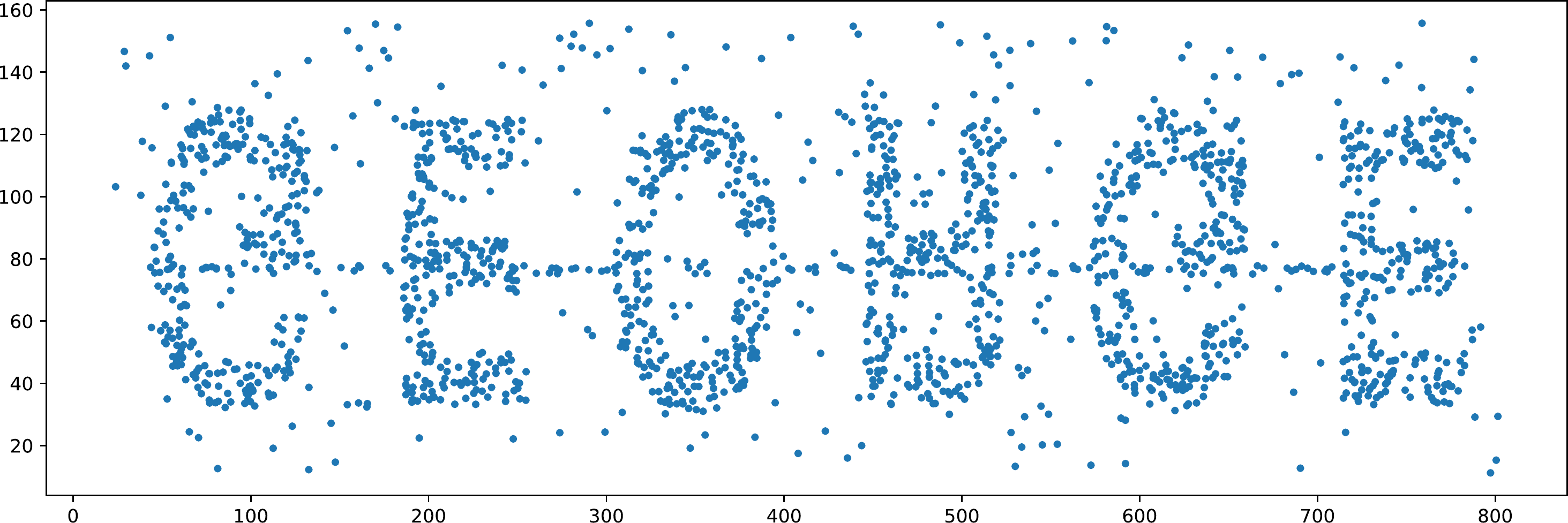} }}
    \\
    \subfloat[\centering 8000 data points ]{{\includegraphics[width=0.3\textwidth]{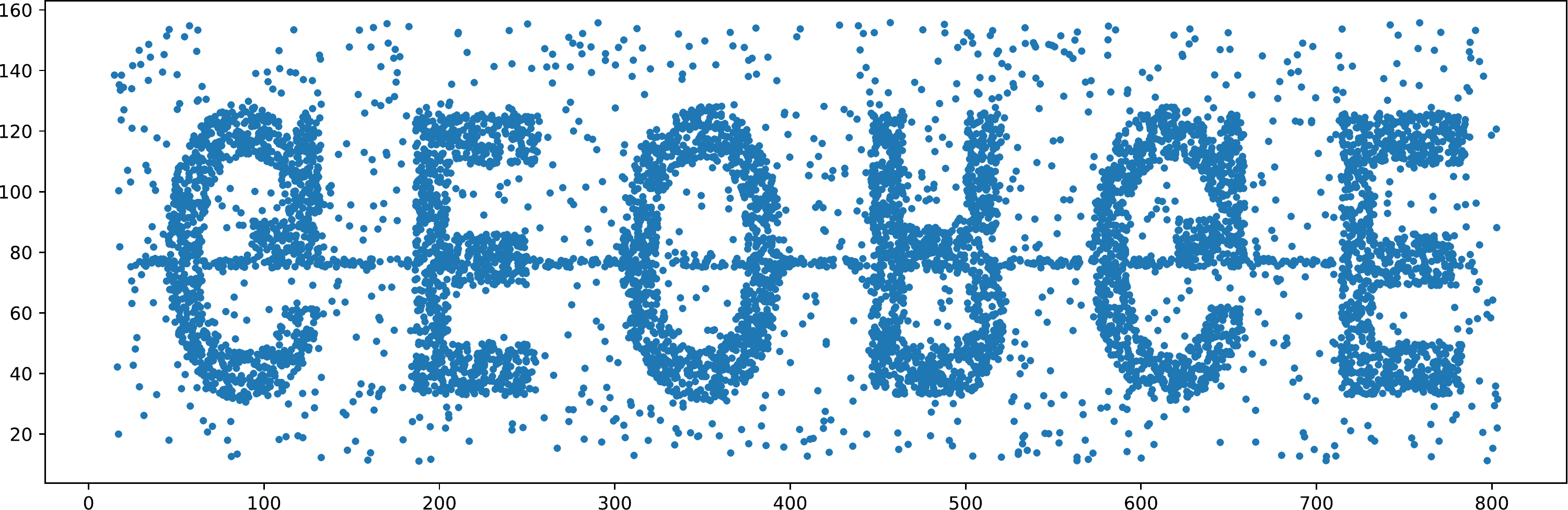} }}
    \caption{Not enough data cannot clarify patterns.}
    \label{fig:9}
\end{figure}

Even a problem has finite domain, we need to have enough data to present its patterns. If the patterns cannot be well presented by a dataset, the ML model cannot learn the correct patterns from the dataset either. Fig. \ref{fig:9} shows an example that the shortage of data could fail to present the patterns.

By the well-known Curse of Dimensionality, the required amount of data to present patterns will increase exponentially in higher dimension space. For 256x256 images, its dimensionality is 65,536. Hence, the training set may not have enough data to train an effective ML model for general counting connected components.

\subsection{Pattern consistency}
\begin{figure}[htbp]
    \centering
    \subfloat[\centering Subset 1 ]{{\includegraphics[width=0.25\textwidth]{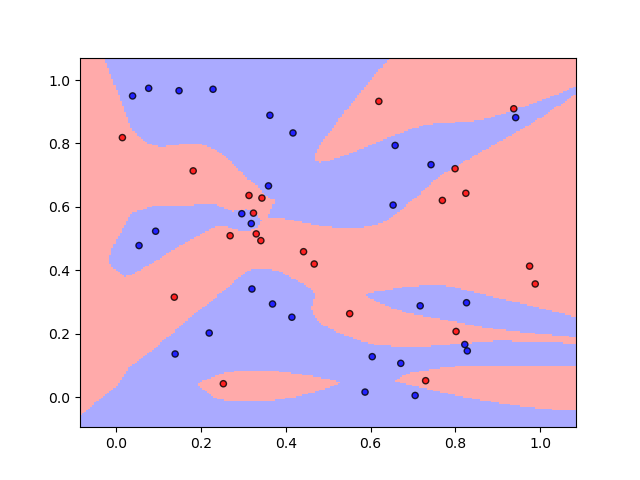} }}
    \subfloat[\centering Subset 2 ]{{\includegraphics[width=0.25\textwidth]{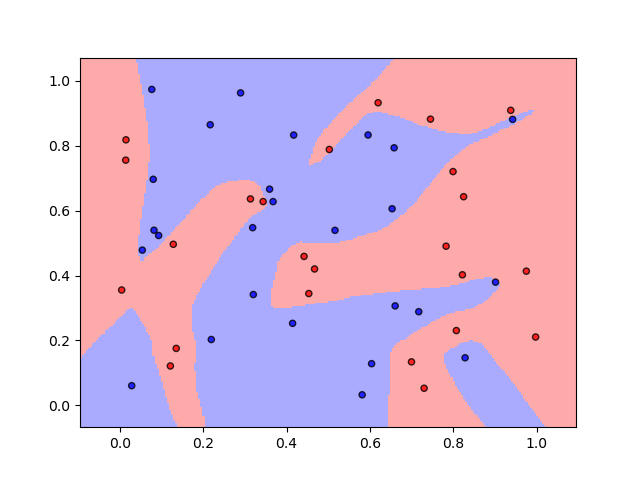} }}
    \caption{Two subsets of random two-class data set and their decision boundaries.}
    \label{fig:10}
\end{figure}

For problems having finite domain, there is another condition that we can never obtain enough data to present its patterns; that is the problem includes infinite number of data and any finite subset does not have the similar pattern as the whole set. In other words, we obviously cannot train ML models on infinite number of data (the whole set) but any finite subset cannot grasp the real pattern of the problem. For example, the infinite random two-class data in $(0,1)^2$. This dataset can have infinite number of data and the decision boundaries of any two subsets of it are different (Fig. \ref{fig:10}). Hence, the infinite random two-class data set is not ML-learnable. Some problems have finite domain and infinite number of data are ML-learnable, such as the Gaussian mixtures models because we could find a subset to present their Gaussian distributions.

For some problems, even they have finite number of data, no subset can describe their patterns unless to use the whole set. For example, the $n$-XOR problems, which to fit $y=x_1 \oplus x_2 \oplus \cdots \oplus x_n;\; x_i \in \{0, 1\} $. It has finite ($2^n$) data and to train models should use all of them.

\begin{figure}[htbp]
\centerline{\includegraphics[width=0.48\textwidth]{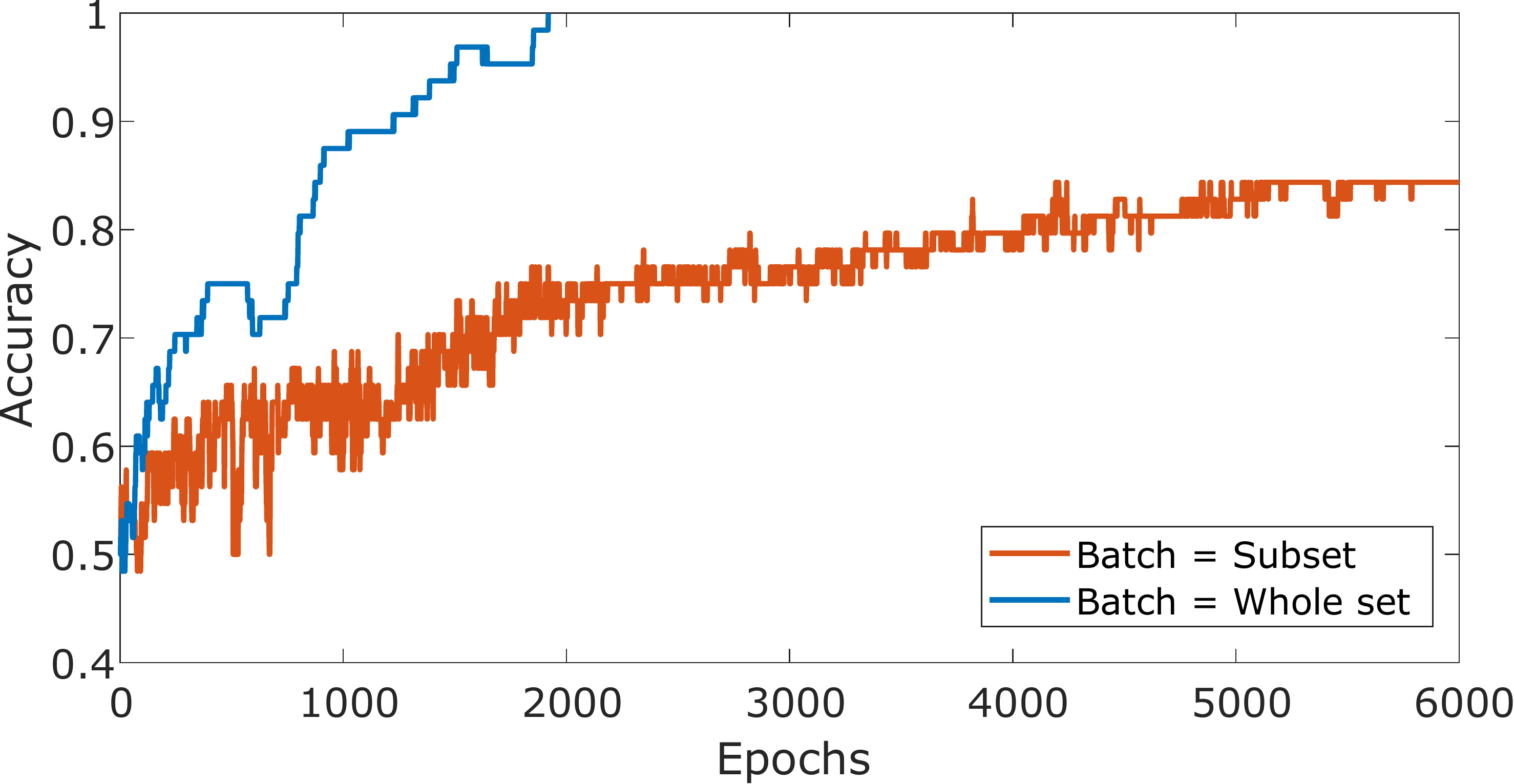}}
\caption{Train 6-XOR dataset by FCNN.}
\label{fig:11}
\end{figure}

We trained a FCNN to fit the 6-XOR problem. The training process in Fig. \ref{fig:11} show that if the training batch is the whole set (batch size $=2^6$), the FCNN model could learn the pattern fast and reach at 100\% training accuracy. But if the training batch is a subset (batch size $=2^6/4$), the training accuracy increases slowly, oscillates frequently, and seems hard to reach at 100\%. Thus, $n$-XOR problems are very difficult to train FCNN models by subsets.

In summary, whatever problems have infinite or finite number of data, if any finite subset does not have the similar pattern as the whole set (no \textit{pattern consistency}), the problems are not ML-learnable.

By examining the cases with no pattern consistency, we could find two key features: 1) the data in dataset are distributed in high density; 2) the mixture of different-class data is complex, which means different-class data are very close to each other and each datum is surrounded by many different-class data. We suppose that the two features are evidence of problems having no pattern consistency.

\begin{figure}[htbp]
\centerline{\includegraphics[width=0.45\textwidth]{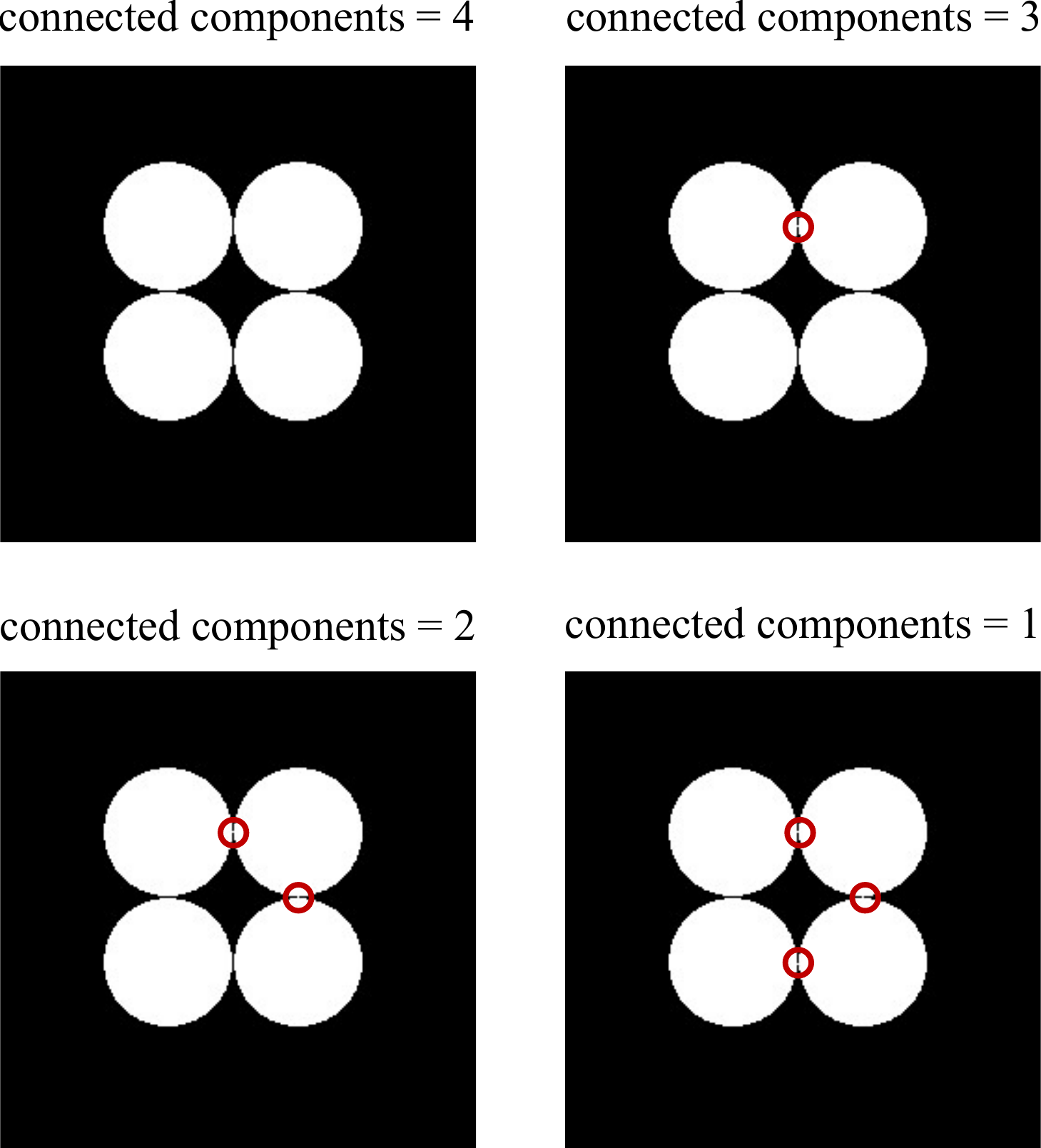}}
\caption{Very similar images with different numbers of connected components. A red ring highlights a white pixel that connects two white circles together to make the two components to become one. Upper left image has four connected components (white circles); upper right image has three connected components because two components (circles) have been connected as one by a pixel; same to the two lower images.}
\label{fig:12}
\end{figure}

Finally, we consider the problems of counting connected components have no pattern consistency because 1) the components could be any number, shapes, sizes, and positions in an image; thus, the dataset is high dense; 2) images with different numbers of connected components can be very similar to each other (close to each other in data space). Fig. \ref{fig:12} shows that one-pixel difference can change the number of connected components. Thus, the mixture of data is complex. Therefore, general counting connected components is not ML-learnable because of no pattern consistency.

\section{Discussions}
Although we have shown that the DNN cannot count the general connected components in images, it can count the objects in images for specific counting problems because they have the three \hyperlink{ml-learn}{ML-learnable characteristics}.

In recent studies that have successfully applied DNNs to object counting, their problems:
\begin{itemize}
    \item have finite domain because they count specific objects, such as persons, cars, leaves, etc. These specific objects have limited patterns domain for shape, size, color, and/or position.
    \item are solved by using pre-processing to reduce dimensionality, such as to apply objects matching, recognition, segmentation, etc. before counting because their objects are specified. Thus, the data they have could be enough to train an effective ML model for counting specific objects.
    \item are not complex and datasets are not high dense because the limitations show before. For example, to count car numbers, images with different numbers of cars are not similar to each other (far from each other in data space).
\end{itemize}

\subsection{Future works}
To show DNNs cannot generally count connected components in images, more experiments in controlled conditions would strengthen this conclusion. And we could provide explanations to these results, such as to answer what leads the counting to fail.

We will refine the theory about pattern consistency to quantitatively define the similar-pattern subsets and find metrics to examine if a problem has such characteristic or not.

In addition, we will try to find the boundary between countable and non-countable problems for DNNs; that is, the boundary between specific and general counting problems in terms of the three ML-learnable characteristics or other applicable theories.
\section{Conclusions}
In this paper, we found DNNs do not have the ability of counting general connected components but can count pixels. We provide many experiments to support our conclusions and explanations to understand the results and phenomena of experiments. We consider that learnable problems for ML models, such as DNNs, should have the three \hyperlink{ml-learn}{ML-learnable characteristics}. The theory of learnable patterns for DNNs has explained why DNNs work for specific counting problems but cannot generally count connected components.

\bibliography{ref}

% \vspace{12pt}
% \color{red}
% IEEE conference templates contain guidance text for composing and formatting conference papers. Please ensure that all template text is removed from your conference paper prior to submission to the conference. Failure to remove the template text from your paper may result in your paper not being published.

\end{document}